\documentclass[conference]{IEEEtran}
\IEEEoverridecommandlockouts
\usepackage{adjustbox} 
\usepackage{array}     
\usepackage{cite}
\usepackage{amsmath,amssymb,amsfonts}
\usepackage{algorithmic}
\usepackage{graphicx}
\usepackage{textcomp}
\usepackage{xcolor}
\def\BibTeX{{\rm B\kern-.05em{\sc i\kern-.025em b}\kern-.08em
    T\kern-.1667em\lower.7ex\hbox{E}\kern-.125emX}}
\begin{document}

\title{Slope-Guided Mamba and Angular-Refined Transformer for Light Field Super-Resolution\\
                           
\thanks{This work was partially supported by the Advanced Materials-National Science and Technology Major Project(No.2024ZD0607800), the National Natural Science Foundation of China (No.62372023), the Zhejiang Provincial Natural Science Foundation of China
(Nos.LQN25F020028, LMS25F020006, LZ24F030012) and the
Research Start-up Funds of Hangzhou International Innovation Institute of Beihang University (Nos.2024KQ087, 2024KQ012).}
}


\author{
\IEEEauthorblockN{Li Jin, Jian Huang, Junde Lu, Shuai Wang, Hao Sheng, Jie Wu$^{\ast}$ \thanks{$^{\ast}$Corresponding author.}}
\IEEEauthorblockA{
Hangzhou International Innovation Institute, Beihang University, Hangzhou, China \\
\textit{\small{\{lijin01, hj, ljd2406107, shuaiwang, shenghao, jiewu\}@buaa.edu.cn}} \\
}
}

\maketitle

\begin{abstract}
Light Field Super-Resolution (LFSR) necessitates accurate modeling of spatial-angular correlations while preserving intrinsic 4D ray coherence. However, maintaining such high-dimensional consistency remains challenging, primarily due to two inherent limitations in prevailing modeling paradigms. First, spatial and angular dimensions are often modeled in a decoupled manner, restricting early cross-dimensional interaction and leading to geometric inconsistencies. 
Moreover, although continuous sequence modeling paradigms show promise in representing epipolar structures, their rigid scanning mechanisms fundamentally conflict with epipolar geometry, limiting geometry-aware feature aggregation.
To address these challenges, we propose a hybrid light field super-resolution network, termed SMART, which integrates a Slope-Guided Mamba and an Angular-Refined Transformer to effectively overcome these limitations.
Specifically, we introduce an angular-modulated spatial module to bridge the decoupling gap, incorporating angular priors to strengthen spatial–angular correlation modeling. To mitigate the scan-geometry mismatch, we propose a manifold-aligned trajectory module that enables geometry-consistent sequence modeling along epipolar structures. 
Experiments on five benchmarks demonstrate that SMART achieves state-of-the-art performance, surpassing previous methods by 0.42 dB (PSNR) with significantly reduced artifacts.
\end{abstract}


\section{Introduction}




Light field imaging records the spatial-angular distribution of light rays, enabling sophisticated applications such as depth estimation and 3D reconstruction \cite{DPT, EPIT}. However, hardware-constrained trade-offs between spatial resolution and angular density typically result in low-resolution sub-aperture images (SAIs), which significantly limit downstream performance. 
Consequently, Light Field Super-Resolution (LFSR) has emerged as a pivotal technique to restore high-frequency details while preserving 4D epipolar geometry, thereby overcoming hardware bottlenecks and enabling high-quality applications such as virtual reality and autonomous driving.
Despite its promise, LFSR is hindered by insufficient spatial-angular modeling, particularly in occluded regions \cite{cong2025seiif}.

To address this core issue, early CNN-based methods (e.g., DistgSSR~\cite{DistgSSR}) employed specific convolution kernels to decouple the 4D light field subspaces. However, due to local receptive fields, they struggle to capture long-range cross-view dependencies. Subsequent transformer-based architectures typically adopt a decoupled processing paradigm for spatial and angular information. For instance, EPIT~\cite{EPIT} employs serial spatial convolution and angular transformers, while LFT~\cite{LFT} utilizes recurrent spatial and angular transformer modules for feature interaction. Although these methods mitigate the deficiencies in long-range context modeling, they primarily rely on shallow feature concatenation mechanisms when integrating the decoupled branches. Consequently, this decoupling paradigm lacks deep geometric awareness and spatial-angular synergy, limiting the ability to achieve global consistency and accurate reconstruction of high-frequency details.

\begin{figure*}[htbp]
  \centering
  \begin{minipage}{0.45\textwidth}
    \centering
    \includegraphics[width=\textwidth, page=1]{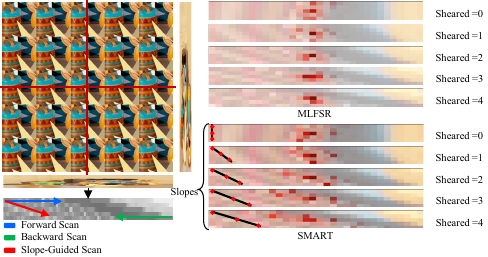}\\
    \vspace{10pt}
     {\small (a)}
  \end{minipage}
  \begin{minipage}{0.43\textwidth}
    \centering
    \includegraphics[width=0.9\textwidth, page=1]{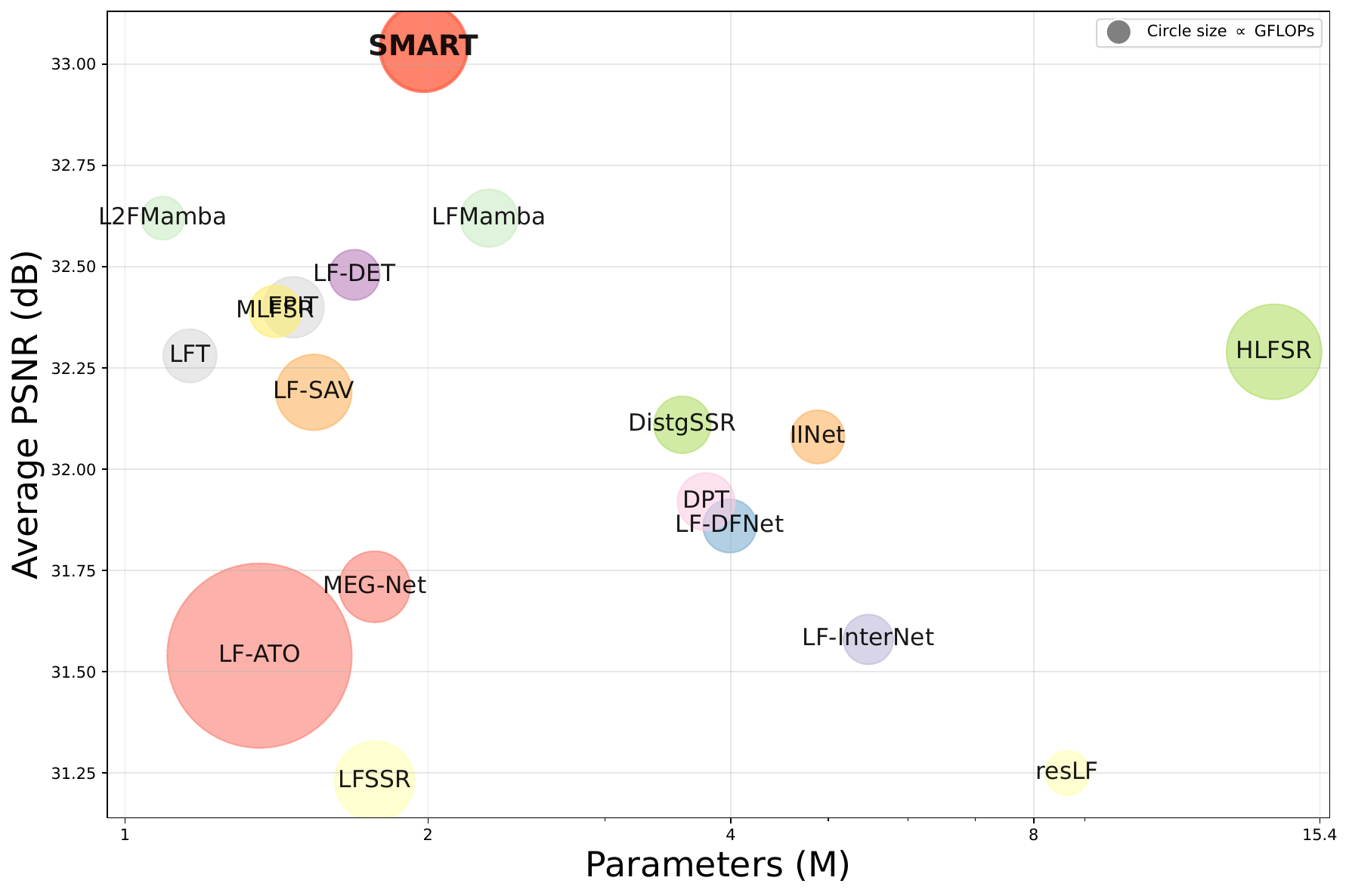}\\
    {\small (b)}
  \end{minipage}
  \caption{Visual analysis and performance comparison. (a) Left: Illustration of the geometric-structural mismatch between the EPI structure and the traditional mamba scanning pattern. Right: Heatmaps comparing traditional mamba and SMART at the same spatial location across different sheared EPIs. (b) Comparison of accuracy and model scales between SMART and state-of-the-art methods.}
  \label{fig:abs}
\end{figure*}

Furthermore, within the realm of LFSR, deep spatial-angular consistency is directly manifested as distinct linear structures in Epipolar Plane Images (EPIs). Leveraging the long-range linearity of EPIs, State Space Models (SSMs)~\cite{liu2024vmamba} offer an efficient sequence modeling framework with linear complexity. However, existing state-of-the-art architectures based on SSMs~\cite{MLFSR,lfmamba} have failed to surmount this geometric representation hurdle. As shown in Fig.~\ref{fig:abs}(a), scene information in the EPI domain manifests as continuous slanted trajectories governed by disparity, rather than being distributed across a regular orthogonal grid. Regrettably, most current SSM-based LFSR methods directly inherit the rigid grid-based scanning patterns from Visual Mamba~\cite{liu2024vmamba}, resulting in a fundamental scanning-geometry mismatch. Specifically, whether referring to MLFSR~\cite{MLFSR}, which introduces bidirectional scanning to capture long-range dependencies, or LFmamba~\cite{lfmamba}, which employs channel-wise lightweight SSMs to facilitate feature fusion, these approaches artificially fragment the physical manifold of EPIs. This isolation of correlated pixels during serialization dilutes information and scatters attention, hindering focus on geometrically consistent epipolar lines and constraining feature extraction precision.


To bridge these gaps, we propose SMART (Slope-Guided Mamba and Angular-Refined Transformer). This hybrid architecture unifies geometry-aligned sequence modeling with structure-aware optimization, synergizing transformer’s texture extraction with mamba’s geometric consistency.

Specifically, to rectify the insufficient spatial-angular synergy inherent in existing decoupled methods, we propose the Angular-Modulated Spatial module (AMS). While leveraging Query-Key based attention to capture intra-view spatial context, this module injects angular priors derived from the macro-pixel layout into the attention maps as structural biases. By incorporating these angular constraints, the model effectively reconstructs high-frequency details in occluded regions, ensuring comprehensive feature completion and geometric consistency.
Simultaneously, we propose the Manifold-Aligned Trajectory  module (MAT), which innovatively employs a slope-guided geometric scanning strategy to address the scan-geometry mismatch. Capitalizing on the unique EPI structure, we explicitly estimate the local slope $k$ governed by disparity and utilize it as a baseline for sampling offsets. This allows the scanning path to adaptively warp, breaking rigid grid constraints to ensure state transitions occur strictly along geometrically consistent epipolar lines rather than irrelevant background pixels. As illustrated in Fig. 1(a), this mechanism aggregates features from the same spatial point across views along their precise geometric trajectories.
Experiments on five standard datasets validate the effectiveness of our approach. As shown in Fig. 1(b), SMART achieves a new SOTA of 33.04 dB / 0.9478 with efficient parameters and GFLOPs, surpassing L$^2$Fmamba \cite{l2fmamba} by 0.42 dB. Qualitative results in Fig. \ref{fig:vis} confirm that SMART restores sharp text outlines and intricate textures while effectively suppressing distortion artifacts.
The contributions can be summarized as follows:
    


(1) We propose SMART, a hybrid network that synergizes the content refinement capabilities of transformers with the geometric consistency modeling of mamba to preserve 4D light field coherence.
(2) We design a slope-guided adaptive scanning mechanism that utilizes estimated EPI slopes $k$ to break rigid grid constraints, enabling mamba to aggregate features along geometrically consistent trajectories.
(3) Extensive evaluations on five benchmark datasets demonstrate that SMART achieves new SOTA results in PSNR and SSIM, significantly outperforming existing methods.

\begin{figure*}[t!] 
    
    \centering 
    
    \includegraphics[width=0.9\textwidth]{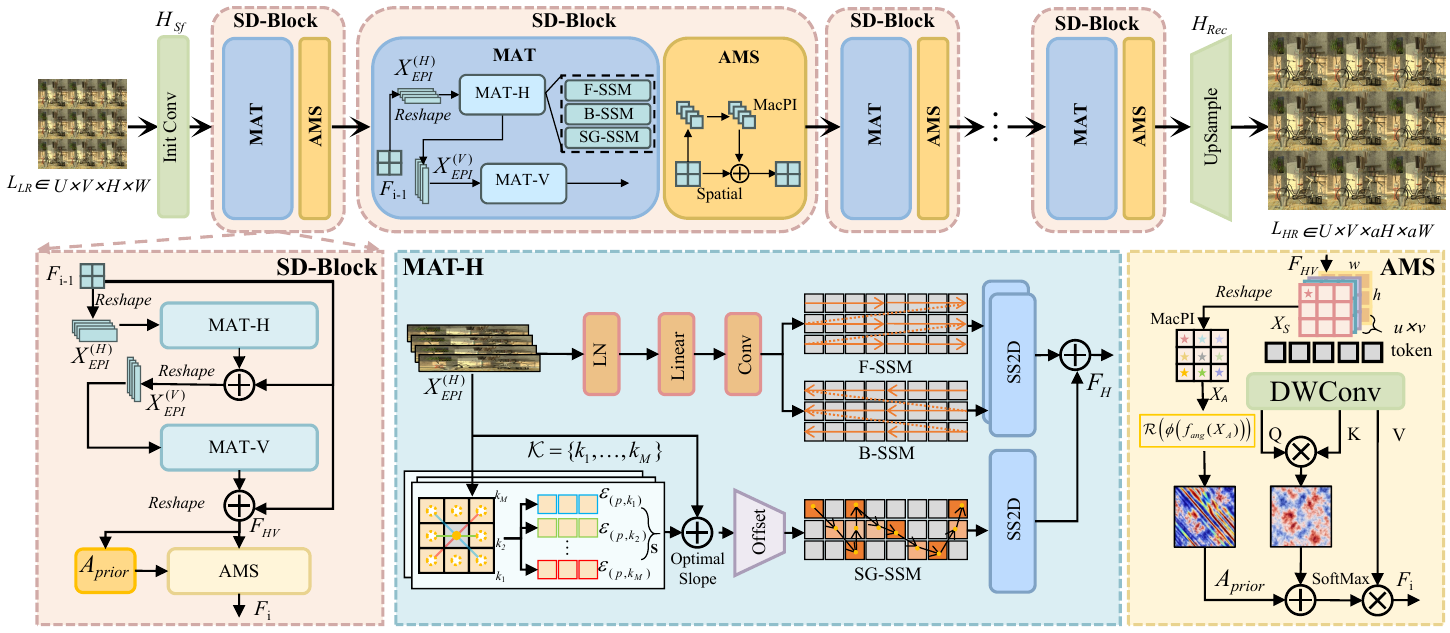}
    
    \caption{The overview of the SMART. The interaction stage consists of $N$ cascaded Structure-Detail blocks (SD-Blocks), where each block comprises a MAT module and an AMS module. The MAT module includes horizontal (H) and vertical (V) EPI processing units, which share a structurally identical design.}

    \label{fig:overall} 
\end{figure*}

\section{Methods}

\subsection{Methodology Overview}
\noindent \textbf{Problem Formulation}. The light field feature is generally represented as a 4D tensor $L \in \mathbb{R}^{U \times V \times H \times W}$, where $(U, V)$ and $(H, W)$ denote the angular and spatial resolutions, respectively. The goal of LFSR is to reconstruct a high-resolution light field $L_{HR} \in \mathbb{R}^{U \times V \times \alpha H \times \alpha W}$ from its low-resolution counterpart $L_{LR} \in \mathbb{R}^{U \times V \times H \times W}$, where $\alpha$ represents the upscaling factor. This problem defines a mapping function $\mathcal{F}$:
\begin{equation}
L_{SR} = \mathcal{F}(L_{LR}; \Theta),
\end{equation}
where $L_{SR}$ denotes the reconstructed LF and $\Theta$ represents the learnable parameters of the network.

\noindent\textbf{SMART Architecture.} To address this problem, we propose SMART, a cascaded network for LFSR. As shown in Fig.\ref{fig:overall}, the pipeline comprises three stages: feature extraction, deep structure interaction, and reconstruction.
\begin{gather}
    F_0 = H_{Sf}(L_{LR}) \label{eq:shallow}, \\
    F_i = AMS_i \left( MAT_i (F_{i-1}) \right), \quad i=1, \dots, N \label{eq:interaction}, \\
    L_{HR} = H_{Rec}(F_{N}), \label{eq:reconstruction}
\end{gather}
where $H_{Sf}$ extracts shallow features from the input $L_{LR}$ using convolution layers. The interaction stage employs MAT module to reinforce epipolar consistency using EPI features, followed by AMS module for disparity and texture encoding. Finally, $H_{Rec}$ upsample $F_{N}$ via sub-pixel convolution to recover the high-resolution $L_{HR}$.



\subsection{Manifold-Aligned Trajectory Module }

Applying SSMs \cite{liu2024vmamba} to LFSR faces a  mismatch between fixed scanning trajectories and the underlying epipolar geometry of the light field.
Specifically, scene points in EPIs manifest as slanted lines rather than orthogonal grid patterns.
Consequently, standard raster scans compel SSMs to model spatially adjacent but geometrically unrelated pixels, disrupting the inherent manifold continuity.
To address this, we propose the MAT module.
It shifts sequence modeling from blind spatial scanning to geometry-aware manifold traversal, guiding feature extraction along continuous epipolar structures.


To leverage the inherent manifold structure, we reorganize {$F_{i-1}$} into a horizontal EPI stack $X_{EPI}^{(H)} \in \mathbb{R}^{(U \cdot H) \times W \times V \times C}$, mapping scene depth to linear slopes in the $V \text{-} W$ plane. We define a disparity hypothesis set $\mathcal{K} = \{k_1, \dots, k_M\}$. For each pixel $p$, we sample features along the trajectory corresponding to $k_m$ within an angular neighborhood $\Omega$. We quantify the degree of consistency among pixels along the trajectory by computing the consistency cost $\mathcal{E}$:
\begin{equation}\mathcal{E}(p, k_m) = \frac{1}{C} \sum_{c=1}^{C} \text{Var} \left( { X_{EPI}^{(H)}(p + \delta \cdot \vec{v}_{k_m}) \mid \delta \in \Omega } \right),\end{equation}
where $\vec{v}_{k_m}$ is the direction vector defined by $k_m$, and $\text{Var}(\cdot)$ computes the trajectory variance. $p$ represents the currently processed pixel point, $C$ denotes the number of feature channels, and $\delta$ is the offset within the angular neighborhood $\Omega$ used for sampling. Lower $\mathcal{E}$ indicates minimal feature dispersion, corresponding to the true scene geometry.

We convert the cost $\mathcal{E}$ into a probability distribution $P$ via Softmax and derive the optimal slope $\mathbf{S}$ as the expectation:
\begin{align}
P(p, k_m) &= \frac{\exp (-\lambda \mathcal{E}(p, k_m))}{\sum_{j=1}^{M} \exp (-\lambda \mathcal{E}(p, k_j))},\\
\small
\mathbf{S}(p) &= \sum_{m=1}^{M} k_m \cdot P(p, k_m)
,\end{align}
where $\lambda$ is a learnable parameter, $\mathbf{S} \in \mathbb{R}^{U \times V\times H \times W}$ encodes dominant EPI directions and provides a continuous and differentiable geometric manifold constraint for feature scanning.

Leveraging the geometric prior $\mathbf{S}$, our slope-guided adaptive scanning mechanism to align SSM receptive fields with the manifold. The input $X_{EPI}^{(H)}$ are flattened into three parallel sequences: forward $X_{fwd}$, backward $X_{bwd}$, and the slope-guided path $X_{sg}$. While the raster paths capture bidirectional context, $X_{sg}$ transcends grid limits, utilizing geometric priors to dynamically reconstruct the scanning trajectory.

The slope prior $\mathbf{S}$ generates coordinate offsets and sorting indices, allowing the SSM to transcend grid limits. This facilitates scanning along epipolar lines, aggregating long-range information from geometrically correlated regions. Specifically, $\mathbf{S}$ is embedded via an MLP and concatenated with feature $F$. This combined input drives an offset network to produce spatial offsets $\Delta P \in \mathbb{R}^{2 \times H \times W}$ and temporal sorting indices $\mathcal{I} \in \mathbb{R}^{1 \times H \times W}$. $[\cdot ; \cdot]$ denotes channel-wise concatenation.
\begin{equation}(\Delta P, \mathcal{I}) = \Psi \left( [\text{MLP}(\mathbf{S}); F_{i-1}] \right),\end{equation}
where $\Psi$ denotes the offset and index generation network. We resample the original features $F_{i-1}$ using $\Delta P$ to achieve feature aggregation along the slope direction, yielding the aligned features $F_{align}$. Subsequently, $F_{align}$ is reordered based on $\mathcal{I}$ to establish the slope-guided dynamic scanning trajectory.

These sequences are processed by a shared selective scan kernel, fused via summation, refined by an MLP, and reshaped, incorporating a residual connection to get the output $F_{H}$:
\begin{equation}
\small
F_{H} = \text{MLP} ( \sum_{d \in {fwd, bwd, sg}} \text{SSM}(X_d) ) + F_{i-1},\end{equation}

where$X_d$ denotes the input sequences. $fwd$, $bwd$, and $sg$ denote the forward, backward, and slope-guided scanning orders, respectively. 
Analogously, for the vertical EPI stack, we transform the features $F_{H}$ into $X_{EPI}^{(V)}$ to obtain the corresponding output $F_V$. The final output of the MAT is the fused representation $F_{HV}$, which combines both $F_H$ and $F_V$.


This design aligns feature extraction with light transport geometry, facilitating consistent high-frequency reconstruction.

\begin{table*}[t]
\centering
\caption{Quantitative comparison (PSNR / SSIM) of different state-of-the-art Light Field Image SR methods for x4 upscaling factor. The last column shows the average results. \textbf{Bold} indicates the best performance, and \underline{underlined} indicates the second best results.}
\label{tab:comparison_results}
\begin{adjustbox}{max width=\linewidth}
\begin{tabular}{lccccccc}
\hline
Methods & \#Params. & EPFL & HCInew & HCIold & INRIA & STFgantry & Average \\
\hline
Bilinear & -- & 24.567 / 0.8158 & 27.085 / 0.8397 & 31.688 / 0.9256 & 26.226 / 0.8757 & 25.203 / 0.8261 & 26.95 / 0.8566 \\
Bicubic & -- & 25.264 / 0.8324 & 27.715 / 0.8517 & 32.576 / 0.9344 & 26.952 / 0.8867 & 26.087 / 0.8452 & 27.72 / 0.8701 \\
\hline
VDSR\cite{vdsr} & 0.67M & 27.246 / 0.8777 & 29.308 / 0.8823 & 34.810 / 0.9515 & 29.186 / 0.9204 & 28.506 / 0.9009 & 29.81 / 0.9066 \\
EDSR\cite{edsr} & 38.89M & 27.833 / 0.8854 & 29.591 / 0.8869 & 35.176 / 0.9536 & 29.656 / 0.9257 & 28.703 / 0.9072 & 30.19 / 0.9118 \\
RCAN\cite{RCAN} & 15.36M & 27.907 / 0.8863 & 29.694 / 0.8886 & 35.359 / 0.9548 & 29.805 / 0.9276 & 29.021 / 0.9131 & 30.36 / 0.9141 \\
resLF\cite{resLF} & 8.65M & 28.260 / 0.9035 & 30.723 / 0.9107 & 36.705 / 0.9682 & 30.338 / 0.9412 & 30.191 / 0.9372 & 31.24 / 0.9322 \\
LFSSR\cite{lfssr} & 1.77M & 28.596 / 0.9118 & 30.928 / 0.9145 & 36.907 / 0.9696 & 30.585 / 0.9467 & 30.570 / 0.9426 & 31.52 / 0.9370 \\
LF-ATO\cite{LF-ATO} & 1.36M & 28.514 / 0.9115 & 30.880 / 0.9135 & 36.999 / 0.9699 & 30.711 / 0.9484 & 30.607 / 0.9430 & 31.54 / 0.9373 \\
LF-InterNet\cite{LF-InterNet} & 5.48M & 28.812 / 0.9162 & 30.961 / 0.9161 & 37.150 / 0.9716 & 30.777 / 0.9491 & 30.365 / 0.9409 & 31.61 / 0.9388 \\
LF-DFnet\cite{LF-DFnet} & 3.99M & 28.774 / 0.9165 & 31.234 / 0.9196 & 37.321 / 0.9718 & 30.826 / 0.9503 & 31.147 / 0.9494 & 31.86 / 0.9415 \\
MEG-Net\cite{MEG-Net} & 1.78M & 28.749 / 0.9160 & 31.103 / 0.9177 & 37.287 / 0.9716 & 30.674 / 0.9490 & 30.771 / 0.9453 & 31.72 / 0.9399 \\
LF-IINet\cite{LF-IINet} & 4.89M & 29.038 / 0.9188 & 31.331 / 0.9208 & 37.620 / 0.9734 & 31.034 / 0.9515 & 31.261 / 0.9502 & 32.06 / 0.9429 \\
DistgSSR\cite{DistgSSR} & 3.58M & 28.992 / 0.9195 & 31.380 / 0.9217 & 37.563 / 0.9732 & 30.994 / 0.9519 & 31.649 / 0.9535 & 32.12 / 0.9440 \\
LFSSR-SAV\cite{LFSSR-SAV} & 1.54M & 29.368 / 0.9223 & 31.450 / 0.9217 & 37.497 / 0.9721 & 31.270 / 0.9531 & 31.362 / 0.9505 & 32.19 / 0.9439 \\
HLFSR-SSR\cite{HLFSR-SSR} & 13.87M & 29.196 / 0.9222 & 31.571 / 0.9238 & 37.776 / 0.9742 & 31.241 / 0.9534 & 31.641 / 0.9537 & 32.28 / 0.9455 \\
\hline
DPT\cite{DPT} & 3.79M & 28.939 / 0.9170 & 31.196 / 0.9188 & 37.412 / 0.9721 & 30.964 / 0.9503 & 31.150 / 0.9488 & 31.93 / 0.9414 \\
LFT\cite{LFT} & 1.16M & 29.255 / 0.9210 & 31.462 / 0.9218 & 37.630 / 0.9735 & 31.205 / 0.9524 & 31.860 / 0.9548 & 32.28 / 0.9447 \\
EPIT\cite{EPIT}& 1.47M & 29.339 / 0.9197 & 31.511 / 0.9231 & 37.677 / 0.9737 & 31.372 / 0.9526 & 32.179 / 0.9571 & 32.42 / 0.9452 \\
LF-DET\cite{LF-DET} & 1.66M & 29.473 / 0.9230 & 31.558 / 0.9235 & 37.843 / 0.9744 & 31.389 / 0.9534 & 32.139 / 0.9573 & 32.48 / 0.9463 \\
\hline
MLFSR\cite{MLFSR} & 1.41M & 29.283 / 0.9218 & 31.562 / 0.9235 & 37.828 / 0.9745 & 31.241 / 0.9531 & 32.029 / 0.9567 & 32.389 / 0.9459 \\
LFmamba\cite{lfmamba} & 2.30M & \underline{29.840} / \textbf{0.9256} & \underline{31.695} / \underline{0.9249} & \underline{37.912}/ \underline{0.9748} & \underline{31.808} /\textbf{ 0.9551} & 31.846 / 0.9533 & 32.621 / 0.9467 \\
L$^2$Fmamba\cite{l2fmamba} & 1.09M & 29.681 / 0.9233 & 31.674 / 0.9243 & 37.864 / 0.9745 & 31.728 / 0.9543 & \underline{32.198} / \underline{0.9574} & \underline{32.623} / \underline{0.9468} \\
\hline
Ours & 1.98M & \textbf{30.284} / \underline{0.9243} & \textbf{31.761} / \textbf{0.9259} & \textbf{38.128} / \textbf{0.9749} & \textbf{32.458 }/ \underline{0.9540} & \textbf{32.572} / \textbf{0.9597} & \textbf{33.040 }/ \textbf{0.9478} \\
\hline
\end{tabular}
\end{adjustbox}
\end{table*}

\begin{table*}[htbp]
  \centering
  \caption{ABLATION STUDY ON THE EFFECTIVENESS OF THE PROPOSED COMPONENTS.}
    \begin{tabular}{ccccccccc} 
    \hline
    AMS & $A_{prior}$ & MAT & EPFL  & HCInew & HCIold & INRIA & STFgantry & Average \\ 
    \hline
          &       &       & 24.567/0.8158 & 27.085/0.8397 & 31.688/0.9256 & 26.226/0.8757 & 25.203/0.8261 & 26.954/0.8566 \\
    $\checkmark$ &       &       & 27.996/0.8897 & 29.835/0.8892 & 35.448/0.9552 & 29.929/0.9293 & 29.401/0.9160 & 30.522/0.9159 \\
 & $\checkmark$ & & 29.491/0.9144& 31.216/0.9194& 37.495/0.9718& 31.846/0.9491& 31.580/0.9520&32.326/0.9414\\
 & & $\checkmark$ & 29.661/0.9161& 31.271/0.9203& 37.426/0.9716& 31.939/0.9498& 31.375/0.9508&32.335/0.9417\\
    $\checkmark$ & $\checkmark$ &       & 29.537/0.9203 & 31.496/0.9214 & 37.777/0.9743 & 31.674/0.9534 & 32.053/0.9559 & 32.507/0.9451 \\
    $\checkmark$ & $\checkmark$ & $\checkmark$ & 30.284/0.9243 & 31.761/0.9259 & 38.128/0.9749 & 32.458/0.9540 & 32.572/0.9597 & 33.041/0.9478 \\
    \hline
    \end{tabular}%
  \label{tab:ablation_study}%
\end{table*}

\subsection{Angular-Modulated Spatial Module }

Existing LFSR methods typically decouple spatial and angular processing, fusing them only at the final stage. This separation ignores the 4D coherence, leading to artifacts caused by geometric inconsistency. 
To address this, we introduce the AMS module, which intertwines spatial with angular geometry by leveraging angular correlations as attention biases.

Specifically, for an input $F_{HV} \in \mathbb{R}^{U \times V \times C \times H \times W}$, we first reshape it into $X_{S} \in \mathbb{R}^{(UV) \times C \times H \times W}$. We employ depth-wise separable convolutions $W_{(\cdot)}$ to project $X_S$ into query ($Q$), key ($K$), and value ($V$) embeddings:
\begin{equation}
Q = W_Q(X_S), \quad K = W_K(X_S), \quad V = W_V(X_S).
\end{equation} 

However, spatial attention derived solely from $Q$ and $K$ is limited to capturing intra-view texture similarities, often overlooking the geometric consistency inherent in light fields. To bridge this gap, we introduce a parallel angular prior branch. Specifically, we reshape the input $F_{HV}$ into macropixels $X_{A} \in \mathbb{R}^{(HW) \times C \times U \times V}$ to explicitly model angular patterns, which are extracted via convolution $f_{ang}$ and compressed by projection $\phi$. Finally, we employ an additive broadcasting operation $\mathcal{R}$ to generate the pairwise attention bias, defined as $B_{i,j} = \phi(x_i) + \phi(x_j)$ for spatial positions $i$ and $j$. The formulation of the angular prior $A_{prior}$ is given by:
\begin{equation}
A_{prior} = \mathcal{R}\left( \phi \left( f_{ang}(X_A) \right) \right).
\end{equation}


The angular prior $A_{prior}$ is integrated into the spatial attention mechanism to impose geometric constraints. We concatenate the spatial affinity $Q \cdot K^T$ with $A_{prior}$ and apply a fusion layer $\mathcal{F}_{mix}$ to generate the attention map:
\begin{equation}
Attn_{map} = \sigma \left( \mathcal{F}_{mix}([Q \cdot K^T, A_{prior}]) \right),
\end{equation}
\begin{equation}
F_{i} = Attn_{map} \odot V + F_{HV},
\end{equation}
where $\sigma$ denotes the activation function and $\odot$ represents element-wise multiplication. This modulation integrates texture similarity with angular consistency, thereby enhancing reconstruction in texture-less regions with depth structures.

\begin{figure*}[t] 
  \centering
  \includegraphics[width=0.9\textwidth]{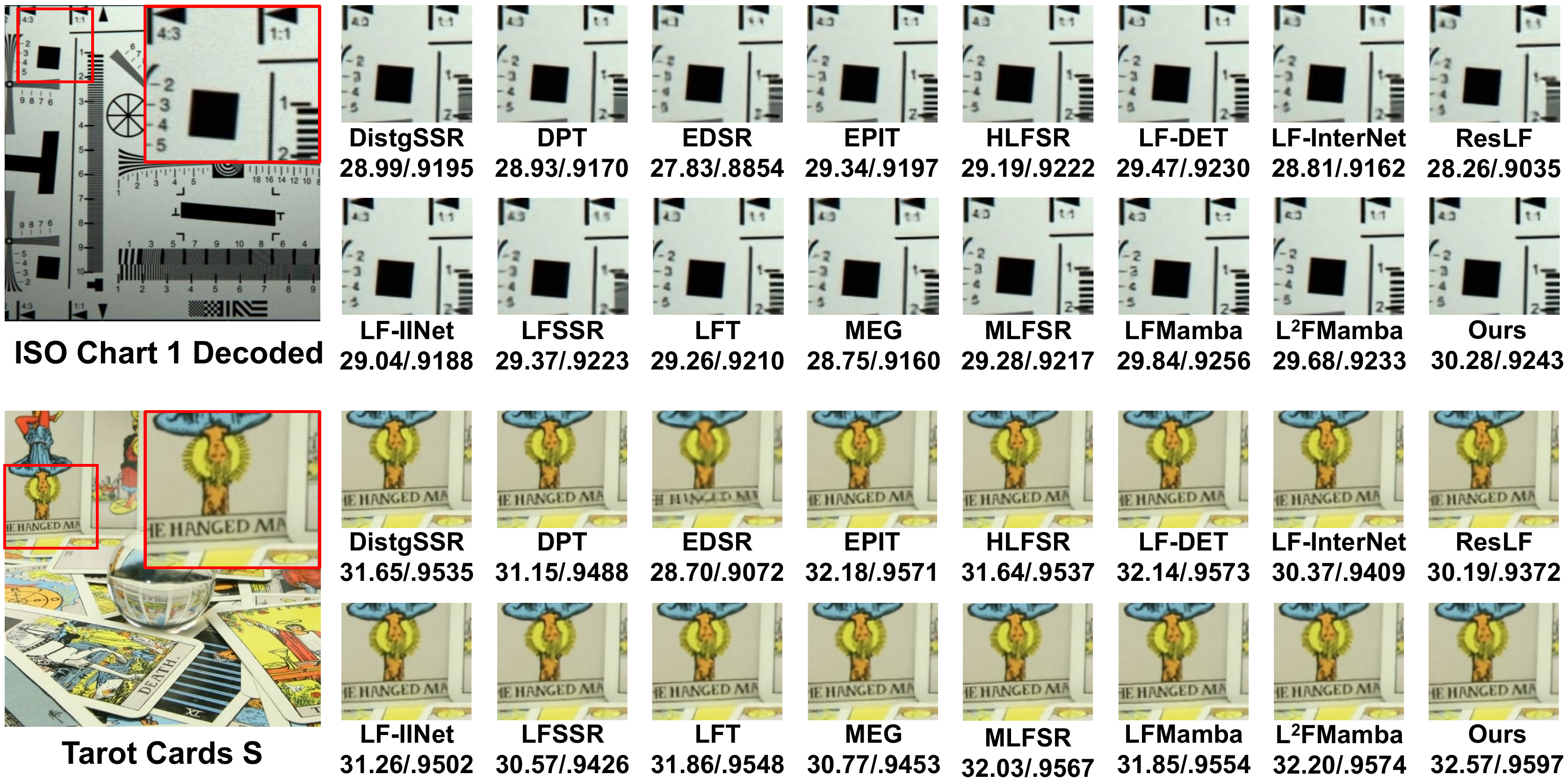} 
  \caption{Visual comparison of $4\times$ super-resolution results on real-world datasets. The first images show the full scene or cropped patches, while subsequent images display the zoomed-in details of the red box regions.}
  \label{fig:vis}
\end{figure*}

\begin{figure}[t] 
  \centering
  \includegraphics[width=0.48\textwidth]{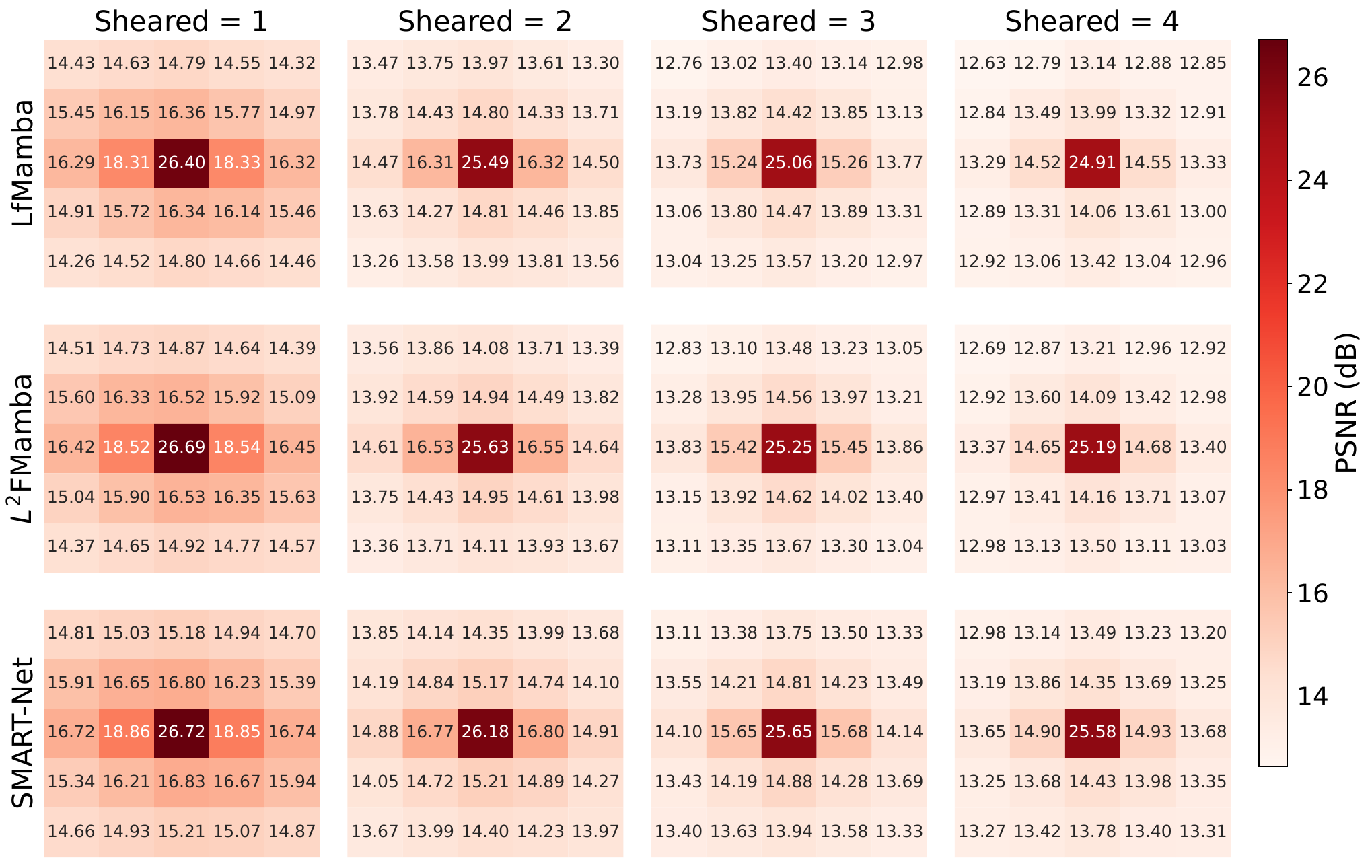} 
  \caption{PSNR distribution across different sheared EPIs on STFgantry.}
  \label{fig:heatmap}
\end{figure}

\section{Experiments}
\subsection{Experimental Settings}
\noindent \textbf{Datasets and Preprocessing.}
We evaluate our method on five publicly available datasets: EPFL \cite{epfl}, INRIA \cite{INRIA}, STFgantry \cite{STFgantry}, HCInew \cite{hclnew}, and HCIold \cite{hclold}. Experiments use the central $5 \times 5$ SAIs, with training patches extracted via a sliding window from high-resolution images and downsampled by a factor of 4 using bicubic interpolation. The proposed model is trained for 100 epochs on 4 NVIDIA GeForce RTX 3090 GPUs. We employ the Adam optimizer with $\beta_1 = 0.9$ and $\beta_2 = 0.999$, and use the L1 loss as our optimization objective. The initial learning rate is set to $2 \times 10^{-4}$ and is decayed by 50\% every 25 epochs. The batch size is set to 2. For the disparity hypothesis set, we initialize $K = \{-2, -1, 0, 1, 2\}$.

\subsection{Comparison with State-of-the-Art Methods}
\noindent \textbf{Quantitative Results.} As summarized in Tab.~\ref{tab:comparison_results}, SMART  exhibits significant superiority over more than 20 representative LFSR methods, achieving an average PSNR of 33.040 dB and an SSIM of 0.9478. With a comparable parameter count, SMART achieves a PSNR gain of 0.417~dB over the previous state-of-the-art method, $L^2$Fmamba. This substantial performance gain validates the exceptional capability of SMART  in modeling complex geometric consistency and reconstructing high-frequency texture details.

\begin{table}[t]
  \centering
  \caption{Ablation Study on the Number of Interaction Blocks}
  \label{tab:stages_ablation}
  \begin{tabular}{ccccc}
    \hline
    Blocks & Params (M) & GFLOPs & Time (ms) & Avg. PSNR/SSIM \\
    \hline
    3 &  1.26&  62.94&35.08 &32.451/0.9428  \\
    4 &  1.61&  82.33&56.65  &32.663/0.9439  \\
    5 &  1.98&  101.72&70.66 &33.040/0.9478  \\
    6 &  2.34&  121.12&86.27  &32.775/0.9459  \\
    \hline
  \end{tabular}
\end{table}

\begin{table}[t]
    \centering
    \caption{Quantitative comparison of different scanning strategies for $4\times$ Light Field Super-Resolution.}
    \label{tab:scanning_strategies} 
    \label{tab:ablation}
    \begin{adjustbox}{max width=\linewidth}
        \begin{tabular}{ccccc} 
            \hline
            Forward & Backward & slope-guided & EPFL & INRIA \\
            (F) & (B) & (SG) & PSNR / SSIM & PSNR / SSIM \\ 
            \hline
            $\checkmark$ & $\checkmark$ &              & 29.584/0.9234 & 31.551/0.9544 \\
            $\checkmark$ &              & $\checkmark$ &   29.949/0.9213         &32.149/0.9521         \\
                         & $\checkmark$ & $\checkmark$ &  
                    30.014/0.9219    &32.251/0.9528              \\
            $\checkmark$ & $\checkmark$ & $\checkmark$ & 30.284/0.9243 & 32.458/0.9540 \\ 
            \hline
        \end{tabular}
    \end{adjustbox}
\end{table}

\noindent \textbf{Qualitative Results.} 
Fig. \ref{fig:vis} illustrates the visual comparison of super-resolution reconstruction results on two real-world datasets. When dealing with fine structures such as text, typical methods, including those based on mamba, often suffer from blurred strokes or distortion artifacts. In contrast, the SMART, by accurately modeling EPI features, is able to restore text outlines, producing reconstruction details that are most consistent with the GT.

\noindent \textbf{Robustness to Scene Geometry.} Fig. \ref{fig:heatmap} illustrates the performance consistency concerning scene geometry on the STFgantry dataset. It demonstrates the PSNR distribution across varying disparity levels. It can be observed that SMART  maintains higher PSNR compared to other methods across disparity range. This confirms that slope-guided adaptive scanning effectively aggregates features along epipolar lines, mitigating performance loss in complex-depth regions.

\subsection{Ablation Study}
\label{sec:ablation}

\noindent \textbf{Component Effectiveness.} 
As shown in Tab.~\ref{tab:ablation_study}, the spatial feature extraction establishes a solid foundation for light field reconstruction, yielding significant PSNR improvements of 2.7--4.2~dB across datasets. The Angle Prior guidance brings further gains of 1.5--2.6~dB, verifying the critical role of angular information in distinguishing complex light field structures. Finally, integrating the slope-guided adaptive
scanning strategy boosts the average PSNR by 0.5~dB. This confirms that this strategy strictly aggregates features along epipolar lines to preserve geometric continuity, achieving more precise reconstruction than generic long-range modeling.

\noindent \textbf{Impact of Cascade Blocks.} 
Tab.~\ref{tab:stages_ablation} shows that as $N$ increases from 3 to 5, the average PSNR rises from $32.451$ dB to $33.040$ dB. Since performance saturates at $N=6$, we select $N=5$ to balance accuracy and efficiency.


\noindent \textbf{Scanning Strategies.} 
As shown in Tab.~\ref{tab:scanning_strategies}, the standard rigid scanning (F+B) yields limited performance. Replacing either fixed direction with the slope-guided boosts performance to 30.014/32.251 dB. The full strategy achieves optimal results, demonstrating that the slope-guided deformable scanning adaptively aligns with light field disparity structures, compensating for the limitations of traditional fixed scanning.

\section{Conclusion}


In this paper, we propose SMART to address the challenge of maintaining high-dimensional geometric consistency in light field super-resolution. SMART incorporates angular priors and epipolar-guided scanning to strengthen spatial-angular correlation and consistent sequence modeling, enabling precise feature aggregation along the intrinsic 4D light field manifold. Experimental results demonstrate that SMART outperforms state-of-the-art methods in reconstruction accuracy, while exhibiting robust capabilities in restoring intricate textures and mitigating geometric inconsistencies.

\bibliographystyle{IEEEbib}
\bibliography{icme2026references}

\end{document}